\documentclass[letterpaper]{article}
\usepackage[preprint]{aaai2027}
\usepackage[hyphens]{url}
\usepackage{graphicx}
\usepackage{pdfpages}
\usepackage{natbib}
\usepackage{caption}
\usepackage{booktabs}
\usepackage{amsmath}
\usepackage{multirow}
\newcommand{\method}{\textsc{OoO-Spec}}
\newcommand{\toolspec}{\textsc{ToolSpec}}

\title{\method: Out-of-Order Semantic Speculation for Fast Tool Calling}
\author{
    Zhiheng Zhang\textsuperscript{\rm 1}\equalcontrib\corresponding,
    Mujie Xu\textsuperscript{\rm 2}\equalcontrib,
    Feiyu Sun\textsuperscript{\rm 3},
    Zhixin Zhang\textsuperscript{\rm 2}
}
\affiliations{
    \textsuperscript{\rm 1}The University of Tokyo\\
    \textsuperscript{\rm 2}Peking University\\
    \textsuperscript{\rm 3}Nanjing University\\
    itsuki-nakano@g.ecc.u-tokyo.ac.jp,
    mujiexu25@stu.pku.edu.cn,\\
    feiyusun@smail.nju.edu.cn,
    zhixinzhang25@stu.pku.edu.cn
}

\begin{document}
\maketitle

\begin{abstract}
LLMs generate tool calls token by token, even though the function choice and argument values can often be predicted in parallel from the request and tool schema. \toolspec{} reduces this cost by drafting schema tokens and retrieving earlier calls, but cannot propose request-specific values absent from either source. We present \method{}, which computes these missing semantics out of order. At request arrival, a Qwen3-0.6B sidecar predicts the function choice and all schema-defined argument slots in one parallel request-level wave while the target begins \toolspec{} decoding. The runtime joins the slot values, renders the resulting call as text, and exposes it to subsequent candidate-construction rounds. The target polls without blocking, re-tokenizes a ready hint with its own tokenizer, and remains the sole verifier and commit authority. The sidecar is trained once with LoRA on Qwen2.5-32B teacher traces and used unchanged across Qwen2.5, Qwen3, and Llama targets, without target-specific drafter training. Across seven fully ranked targets and three benchmarks under greedy batch-one decoding, \method{} is fastest among all evaluated methods in all 21 target--benchmark cells, reaching $2.46\times$--$5.34\times$ over autoregressive decoding with an unweighted mean of $3.89\times$, versus $2.95\times$ for \toolspec{}. It also outperforms every evaluated released learned drafter in each comparable cell. Across Qwen3-4B, 8B, 14B, and 32B targets, the same sidecar improves on \toolspec{} by 34.1\% on average. Its compact semantic payload averages 85 bytes per request excluding protocol metadata, supporting effective split-GPU overlap.
\end{abstract}

\section{Introduction}

Large language models are frequently used for \textbf{tool calling}, repeatedly generating strict, often lengthy function calls. Although each call is a structured object, the decoder treats it as a flat token sequence. Autoregressive decoding therefore requires a forward pass of the target model for every token in its function name, field names, delimiters, and argument values. Yet the function and many argument values can often be inferred directly from the request and schema, well before the decoder reaches their positions in the output. Thus, the target generates the call sequentially even though much of its content can be computed earlier or in parallel.

\begin{figure}[t]
    \centering
    \includegraphics[width=0.88\columnwidth]{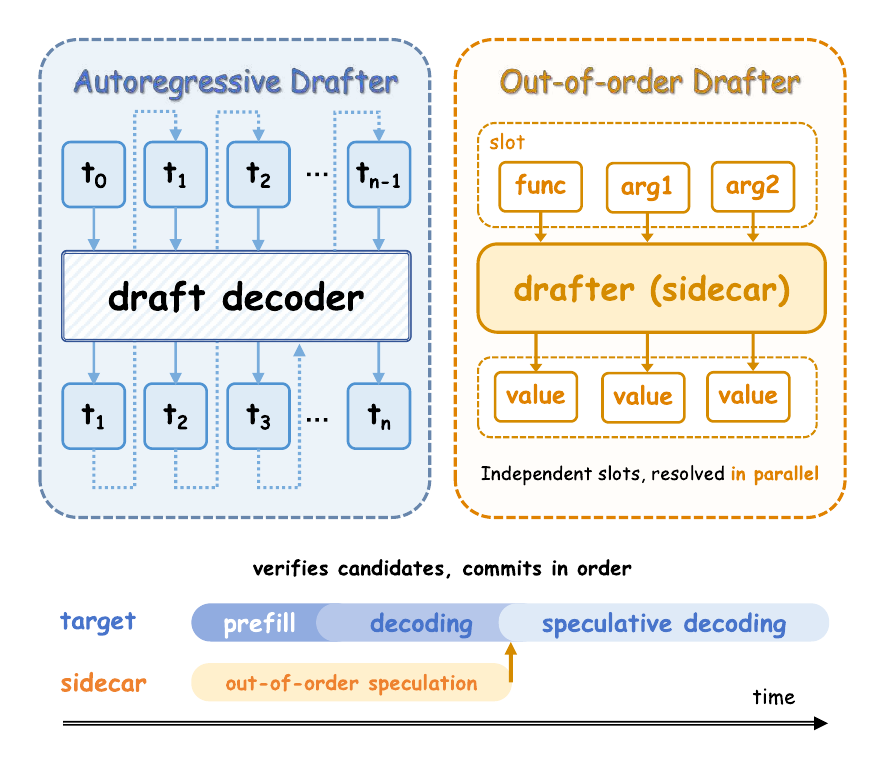}
    \caption{Autoregressive drafting versus out-of-order speculation. Top: an autoregressive drafter proposes tokens one at a time, while an out-of-order drafter resolves independent slots in parallel. Bottom: the sidecar runs alongside the target timeline and sends its result as a hint that joins speculative decoding as additional draft candidates.}
    \label{fig:semantic-slots}
\end{figure}

\textbf{Speculative decoding} reduces this serial cost by proposing multiple future tokens and verifying them together with the target \citep{leviathan2023fast,chen2023accelerating}. Learned drafters such as EAGLE-3, PARD-2, and DFlash can produce strong proposals, but their checkpoints or proposal interfaces are closely tied to a target model \citep{li2025eagle3,an2026pard2,chen2026dflash}. Other methods avoid training a target-specific drafter by reusing tokens from the prompt, generation history, or schema \citep{saxena2023prompt,luo2025token,hu2024sam}. \toolspec{} builds specifically on two properties of tool calling: most output tokens are fixed by the schema, and similar calls recur across requests \citep{xia2026toolspec}. It uses a finite-state machine (FSM) to alternate between schema-token filling and speculative generation for variable fields, and retrieves similar completed calls as additional draft candidates. The target packs these candidates into a tree and verifies them over repeated decoding rounds.

However, \toolspec{} can only reuse what already exists: schema tokens and values retrieved from History Calls. Values new to the current request still fall back to autoregressive decoding. Addressing this gap raises three challenges. \textbf{(1) Generate unseen values.} Schema filling provides syntax, while retrieval only reuses earlier calls. A proposer must infer new request-specific values. \textbf{(2) Generate in parallel.} As we observed, semantic fields need not be computed in textual order. A proposer must run in parallel without blocking the target, so that its latency does not enter the decoding path. \textbf{(3) Use whenever ready.} A concurrent proposal becomes ready at an unpredictable point in the target's decoding, and the target never waits for it. To make full use of concurrency, the system must be able to use a proposal no matter when it becomes ready.


Based on these observations and challenges, we introduce \method{}, an asynchronous semantic speculation system for tool calls. As Figure~\ref{fig:semantic-slots} shows, a lightweight sidecar resolves the function and argument values as independent slots, in parallel and out of textual order, while the target immediately begins the decoding loop. Once ready, the sidecar's outputs join a later candidate-construction boundary as a current-request semantic hint. The target never waits for the hint and remains responsible for verifying and committing every token.

Our sidecar is Qwen3-0.6B \citep{yang2025qwen3} with one LoRA adapter trained on Qwen2.5-32B teacher traces \citep{qwen2024qwen25,hu2022lora}. We reuse this sidecar---base draft model and adapter together---unchanged across all targets and workloads. This portability applies to the learned proposer; each target keeps its own \toolspec{} integration, tokenizer, and frozen rendering policy. The sidecar returns only request-level semantic strings, so it can serve every target and run asynchronously on a separate device.

We evaluate seven targets from the Qwen2.5, Qwen3, and Llama families on API-Bank, ToolAlpaca, and BFCL under greedy batch-one decoding \citep{li2023apibank,tang2023toolalpaca,patil2025bfcl,dubey2024llama3}. Across all 21 target--workload combinations, \method{} is the fastest applicable method and reaches up to $5.34\times$ speedup over autoregressive decoding; its mean speedup is $3.89\times$, compared with $2.95\times$ for \toolspec{}. Against the released learned drafters---EAGLE-3, PARD-2, and DFlash---\method{} leads in every comparable cell across Llama-3.1-8B and Qwen3-4B/8B/14B. With the same sidecar, the gain over \toolspec{} does not shrink as the target grows: across Qwen3 4B, 8B, 14B, and 32B, \method{}'s per-target overall speedup exceeds \toolspec{}'s by 27.1\%, 31.0\%, 40.9\%, and 37.3\%, averaging 34.1\%.

Our contributions are:
\begin{itemize}
    \item We introduce Out-of-Order semantic Speculation for tool calls: a sidecar computes schema-defined function and argument values ahead of their textual positions while the target continues to commit tokens in order.
    \item We make the sidecar's proposal usable whenever it becomes ready: every candidate-construction boundary in \toolspec{} is a join opportunity, so a proposal that arrives mid-generation still accelerates the tokens not yet committed.
    \item We show that one Qwen3-0.6B sidecar transfers across target sizes, model families, and three tool-use workloads without per-target drafter training, with sidecar and communication costs counted in end-to-end time.
\end{itemize}

\section{Related Work}

\paragraph{Learned speculative decoding.}
Classical speculative decoding uses a smaller autoregressive model to draft a token continuation that is subsequently verified by the target \citep{leviathan2023fast,chen2023accelerating}. Later methods reduce the serial depth of drafting. Medusa predicts several future positions with additional decoding heads, while EAGLE and EAGLE-3 construct proposals from target-aligned features \citep{cai2024medusa,li2024eagle,li2025eagle3}. PARD-2 and DFlash further parallelize block drafting with target-aligned or diffusion-style drafters \citep{an2026pard2,chen2026dflash}, and semantic-aware variants probe the target's internal states \citep{dong2026semanticspec}, again bound to one target's representations. All of them, however, require training a dedicated drafter for every target model they serve, at substantial cost. In contrast, \method{} drafts schema-defined semantic strings, allowing one frozen sidecar to serve multiple targets without target-specific retraining.

\paragraph{Train-free and retrieval-based drafting.}
Another line of work avoids training a target-specific drafter by reusing tokens already available at inference time. Prompt Lookup Decoding retrieves repeated prompt spans, Token Recycling builds candidates from observed token transitions, and SAM-Decoding uses a suffix automaton to retrieve continuations \citep{saxena2023prompt,luo2025token,hu2024sam}. \toolspec{} specializes this approach to function calls through schema-driven drafting and History-Calls retrieval \citep{xia2026toolspec}, described in Section~\ref{sec:preliminaries}. Effective as these sources are, none of them can supply a request-specific value that has never appeared in the schema, the context, or the history.

\section{Preliminaries}
\label{sec:preliminaries}

\begin{figure*}[t]
\centering
\includegraphics[width=\textwidth]{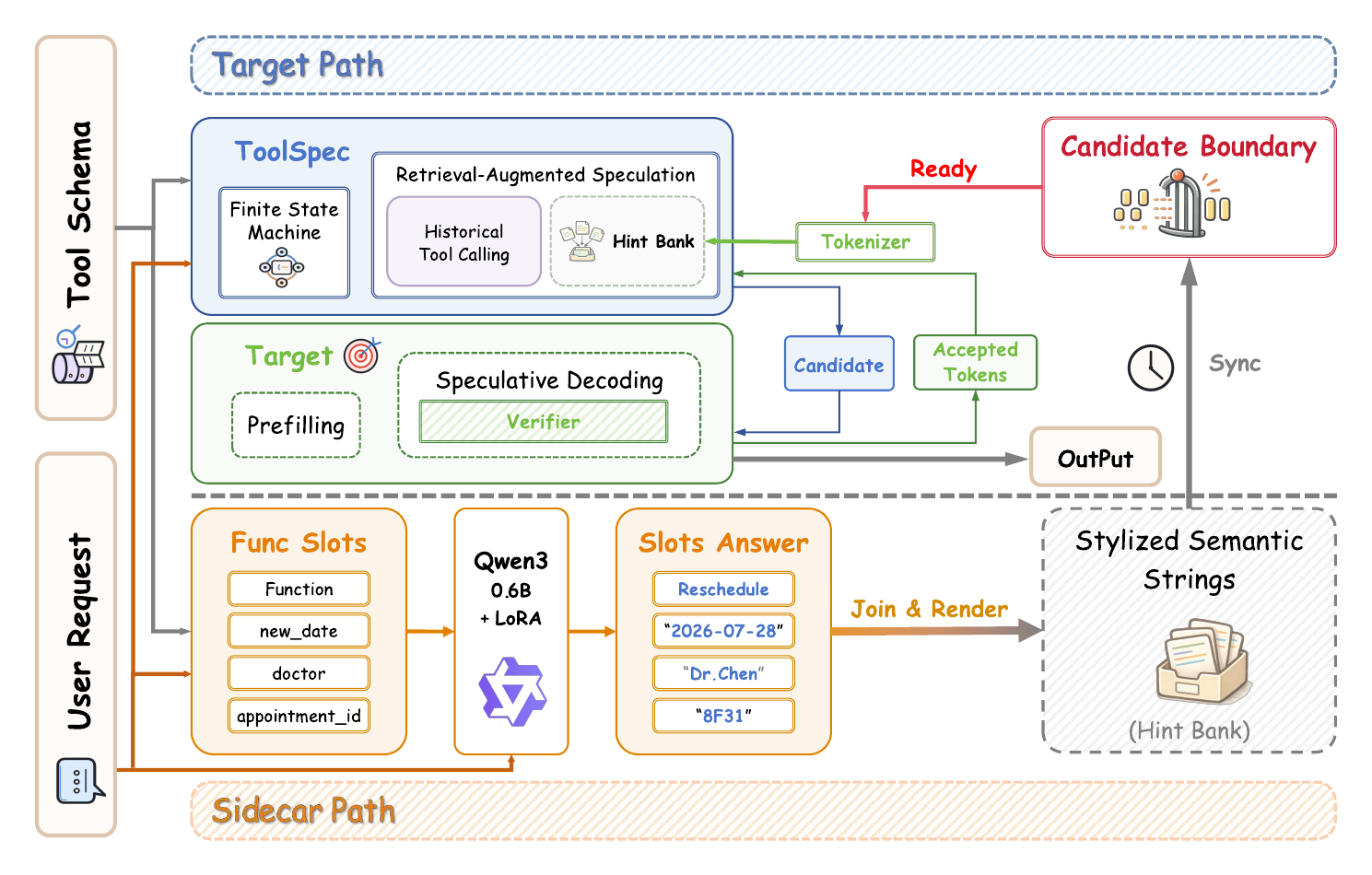}
\caption{Online inference in \method{}. The sidecar's slot predictions are joined and rendered into a hint bank that enters \toolspec{}'s retrieval at candidate-construction boundaries without blocking the target; the target verifier alone commits output tokens.}
\label{fig:overview}
\end{figure*}

\paragraph{Target authority.}
Let $x$ be a request and $y_{<t}$ the prefix already committed by the target.
Under greedy decoding, the next committed token is
\begin{equation}
    y_t=\arg\max_v p_T(v\mid x,y_{<t}).
\end{equation}
A speculative method changes which continuations are evaluated together,
affecting the number of tokens accepted per step and the resulting latency.
Only the target extends $y_{<t}$ and determines the output.

\paragraph{The \toolspec{} decoding loop.}
\toolspec{} specializes target-verified speculation to function calls
\citep{xia2026toolspec}. It maintains two proposal sources. A finite-state
machine compiled from the tool schemas tracks the current region of the call
and supplies constrained scaffolding such as field names and delimiters.
History Calls provide retrieved continuations for variable regions. At each
round, \toolspec{} combines both sources into a candidate tree and evaluates
it with the target. The target commits the longest approved prefix. If decoding
continues, \toolspec{} builds the next tree from all tokens committed so far.

We call each point at which a new tree is built a
\emph{candidate-construction boundary}. These boundaries recur throughout a
call. A later round can therefore include candidates that were unavailable
when decoding began, while tokens already committed by the target remain
unchanged. Our method uses this recurring construction step to add one
candidate source whose proposals may arrive after decoding starts. \toolspec{}
continues to control the FSM, History Calls, candidate budget, packed
verification, KV-cache updates, and recovery.

\section{\method}
\label{sec:method}

\method{} consists of a shared semantic sidecar and an asynchronous inference
loop that connects it to the target. The sidecar is trained once and reused
across target models. At inference time, it predicts the function choice and
all schema-defined argument slots in a single parallel batch, concurrently with
the target's \toolspec{} loop. The target checks the resulting hint at eligible
candidate-construction boundaries without blocking and remains responsible for
verification and left-to-right token commitment. We first describe sidecar
training and then the two online inference paths.

\subsection{Training a Shared Semantic Sidecar}

The supplementary material diagrams the offline training pipeline and lists the frozen asset identities.

\paragraph{Training requests.}
We construct fixed training and development splits from API-Bank and
ToolAlpaca \citep{li2023apibank,tang2023toolalpaca}: 6{,}200 training and 335
development requests from API-Bank, and 3{,}462 training and 194 development
requests drawn only from ToolAlpaca's training file. ToolAlpaca is
decontaminated by removing an entire API whenever its API name or any function
name appears in the evaluation inventory, and we use neither its golden answers
nor its tool-execution outputs. Both datasets use a deterministic prompt-hash
split, giving 9{,}662 training and 529 development source requests with zero
request-ID or prompt-hash overlap between them.

\paragraph{Teacher traces.}
We fix Qwen2.5-32B-Instruct as the offline teacher \citep{qwen2024qwen25}. It
receives each dialogue together with its tool schemas and generates a
structured call greedily. We parse supported tool-call surface forms into a
strict-JSON domain and retain the first valid call, keeping only the normalized
function name and argument map; teacher token IDs, chat-template tokens, and
tool-call control tokens are discarded.

\paragraph{Slot supervision.}
Each retained semantic call is expanded into three task types under the
Qwen3-0.6B chat template. A \emph{function-index} row asks for the selected
function's local index in the current schema. An
\emph{argument-value-or-null} row is created for every parameter of every
candidate function; its target is the compact JSON value when the parameter
belongs to the selected function and appears in the teacher call, and
\texttt{null} otherwise. A third, auxiliary \emph{direct-call} row asks for the
complete normalized call. The expansion produces 75{,}567 training and 3{,}239
development rows. The online sidecar queries only function-index and argument
rows.

\paragraph{Optimization and freezing.}
We initialize an unadapted Qwen3-0.6B \citep{yang2025qwen3} and train one LoRA
adapter \citep{hu2022lora} for a single epoch, masking prompt labels so the
objective is causal cross-entropy over completion tokens; the supplementary
material lists the full hyperparameters. After the fixed development pass we
freeze the epoch-one adapter. The same Qwen3-0.6B base and adapter serve every
target and workload, so deployment to a new target requires no target-specific
gradient update, and the teacher is absent at inference time.

\subsection{Asynchronous Main Inference Loop}

With the shared adapter frozen, online inference follows the two paths in
Figure~\ref{fig:overview}. At request arrival, the runtime submits the complete
sidecar job to a background future and immediately starts target prefill. In
the reported split configuration, the target and sidecar run as separate
processes on two GPUs, and decoded slot strings travel over loopback HTTP;
colocated execution is used only as an ablation. One batched sidecar wave is
launched per request, whereas the target repeatedly constructs and verifies
candidate trees. A completed future can affect candidate construction only at
boundaries where \toolspec{} invokes its retrieval finder. These checks observe
availability and never stall target execution.

We use \emph{sidecar path} to include the background response processing that
follows model generation. The learned sidecar itself produces only slot
answers. Parsing, deterministic assembly, rendering, and target tokenization
remain runtime operations.

\subsection{Sidecar Path: From Schemas to a Hint Bank}

\paragraph{Slot construction.}
Let $\mathcal{F}$ be the candidate functions in a request and $P_f$ the
parameters of function $f$. The runtime constructs the slot set
\begin{equation}
\begin{aligned}
\mathcal{S}&=\{R\}\cup
\{A_{f,p}:f\in\mathcal{F},\ p\in P_f\},\\
|\mathcal{S}|&=1+\sum_{f\in\mathcal{F}}|P_f|.
\end{aligned}
\end{equation}
$R$ asks for the local function index. Each $A_{f,p}$ identifies one candidate
function and parameter and asks for a compact JSON value or \texttt{null}.
Every prompt contains the dialogue and indexed schemas. In particular, an
argument prompt does not receive the answer to $R$.

\paragraph{Batched prediction.}
All slots are submitted together as one request-level vLLM \citep{kwon2023vllm} wave, with greedy
generation and at most 32 new tokens per slot. Waiting for $R$ before issuing
the selected function's arguments would add a serial stage. Instead, argument
slots for unselected functions are computed speculatively alongside the
selected branch.

\paragraph{Join and render.}
After all slot answers return, the response path extracts the first decodable
JSON value from each answer. It maps the function value to a legal local index
under a deterministic fallback, keeps non-null argument values for the
selected function, and discards unavailable or malformed values. The runtime
then joins the function name and remaining values with schema-supplied field
names and JSON structure to form one normalized object. An object or array
parameter remains one JSON-valued slot, and the online path does not issue a
separate whole-call prompt. The normalized object is rendered under a small
frozen policy into equivalent textual views. Together, these views represent
one semantic prediction, not multiple predicted calls. Qwen3 targets use fixed
plain-JSON, Markdown, and XML-style views. For Qwen2.5 and Llama targets, the
layout policy is frozen from training and development traces before formal
evaluation. This policy is target-specific configuration and requires no
gradient update. Raw sidecar token IDs are discarded, so the device boundary
carries decoded slot strings and the target retains control of its own
tokenization.

\subsection{Target Path: Prefill and Nonblocking Hint Pickup}

\paragraph{Prefill.}
The target begins prefill as soon as the request arrives and then enters its
native \toolspec{} loop. Schema-FSM rounds remain unchanged regardless of the
sidecar state. While the future is pending, retrieval rounds use the original
History Calls and fallback candidates.

\paragraph{Candidate boundary.}
At a retrieval-candidate round $t$, the runtime performs a nonblocking
readiness check and forms
\begin{equation}
\mathcal{C}_t=
\begin{cases}
\mathcal{C}^{\mathrm{TS}}_t,
    & \text{sidecar future pending},\\
\operatorname{Merge}_{B}
    (\mathcal{C}^{\mathrm{hint}}_t,\mathcal{C}^{\mathrm{TS}}_t),
    & \text{hint bank ready},
\end{cases}
\end{equation}
where $\mathcal{C}^{\mathrm{TS}}_t$ denotes the native \toolspec{} candidates
for that retrieval round and $\operatorname{Merge}_{B}$ preserves its fixed
candidate budget. A future that completes just after one check is first
observed at a later retrieval boundary; the target performs no rollback.

\paragraph{Tokenizer and hint bank.}
The target-side bridge tokenizes each rendered view once, inserts the target
EOS token between views, and concatenates them into one request-level hint
bank. Encoding and device materialization occur once, and later retrieval
rounds reuse the same bank. Readiness is queried only at retrieval-candidate
rounds.

\paragraph{Suffix matching.}
To retrieve a continuation for the target's current position, let
$s_t=\operatorname{suffix}_n(y_{<t})$ be the length-$n$ suffix of its committed
prefix. The finder searches the encoded bank for exact occurrences of $s_t$,
trying $n=7$ down to $1$, and proposes the tokens following a match. Matching
the current suffix prevents a late hint from proposing content for a position
that the target has already committed. Hint continuations fill the available
\toolspec{} lanes first, and its native sources fill the remaining lanes. If
the bank has no aligned continuation, the round uses only native candidates.
Repeating this alignment at later retrieval boundaries also lets one
request-level bank remain useful after an earlier field has already been
generated or rejected.

\subsection{Target-Only Verification and Recovery}

\paragraph{Verify and commit.}
Hint continuations and native \toolspec{} continuations are compiled into the
same candidate tree. The target evaluates the tree in one forward pass and
commits the longest prefix allowed by its greedy decision rule. At the first
mismatch, it rejects the remaining candidate suffix and uses its own next
token to continue decoding. Only states on the accepted target path enter the
committed KV cache; sidecar outputs never update that cache or commit tokens.

\paragraph{Recovery.}
Rejecting one continuation leaves the rest of the request-level bank
available for later suffix matches. An inaccurate hint can consume candidate
or verification work, and a hint that finishes after target completion is
ignored. In every case, the target determines the committed output. \method{}
changes candidate source and availability; \toolspec{} and the target retain
candidate-tree construction, verification, cache updates, and recovery.

\section{Experimental Setup}
\label{sec:setup}

\subsection{Targets, Workloads, and Protocol}

The fully ranked comparison uses seven targets: Qwen2.5-7B/14B \citep{qwen2024qwen25}, Llama-3.2-3B and Llama-3.1-8B \citep{dubey2024llama3}, and Qwen3-4B/8B/14B \citep{yang2025qwen3}; Qwen3-32B and Qwen2.5-32B join the scaling study. Every method sees the same 597 API-Bank \citep{li2023apibank}, 193 ToolAlpaca \citep{tang2023toolalpaca}, and 68 BFCL Java/JavaScript \citep{patil2025bfcl} requests; the BFCL set is drawn from the v4 Java and JavaScript categories. Table~\ref{tab:datasets} summarizes the three evaluation sets. Tool repetition falls from 8.91 on API-Bank to 1.00 on BFCL, where every request exposes a distinct tool, so history retrieval has the least to reuse there.

\begin{table}[t]
\centering
\small
\begin{tabular}{lrrr}
\toprule
Benchmark & \#Tools & \#Requests & Avg. Tool Rep. \\
\midrule
API-Bank & 70 & 597 & 8.91 \\
ToolAlpaca & 94 & 193 & 2.99 \\
BFCL (Java/JS) & 68 & 68 & 1.00 \\
\bottomrule
\end{tabular}
\caption{Evaluation-set statistics. \#Tools counts distinct tools in the request prompts; Avg.\ Tool Rep.\ is golden invocations per invoked tool, following \citet{xia2026toolspec}.}
\label{tab:datasets}
\end{table}

All measurements use greedy decoding, batch size one, and H100-80GB GPUs, with generation limits fixed across methods per benchmark. Qwen2.5 and Llama targets run \toolspec{}'s FP16 cache backend; Qwen3 targets use the architecture-specific BF16 Transformers \citep{wolf2020transformers} port. \method{} always places the target and the sidecar on two separate GPUs. Reported request latency is wall time from before draft submission until both target and sidecar have finished, so sidecar launch, communication, and coordination are all inside the measurement.

\subsection{Baselines and Placement}

We compare live AR (Vanilla), Prompt Lookup Decoding (PLD), Token Recycling (TR), SAM-Decoding (SAMD), and \toolspec{} on every target, each through its native proposal and verification path with its published configuration. Where an author-supported checkpoint exists we add EAGLE-3, PARD-2, and DFlash; each learned drafter uses its faster measured eligible placement (EAGLE-3 split; PARD-2 and DFlash their official colocated Transformers/Transformers+ paths).

\subsection{Metrics and Correctness Audit}

Following \toolspec{}, each cell reports mean accepted tokens per target verification call (\#MAT), mean per-request tokens per second, and elapsed-time speedup against the live AR reference for the same target and workload; the overall column is the unweighted mean across the three benchmarks. Speed is interpreted only after correctness: for every request we compare the method's output tokens with the live greedy AR trajectory of the same runner.

\section{Results}
\label{sec:results}

\begin{table*}[tp]
\centering
\begingroup
\setlength{\tabcolsep}{2.45pt}
\renewcommand{\arraystretch}{0.93}
\scriptsize
\resizebox*{!}{\dimexpr\textheight-66pt\relax}{%
\begin{tabular}{llrrrrrrrrrr}
\toprule
\multirow{2}{*}{Target} & \multirow{2}{*}{Method} & \multicolumn{3}{c}{API-Bank} & \multicolumn{3}{c}{ToolAlpaca} & \multicolumn{3}{c}{BFCL Java/JS} & \multirow{2}{*}{Overall $\times$} \\
\cmidrule(lr){3-5}\cmidrule(lr){6-8}\cmidrule(lr){9-11}
 & & \#MAT & Tok./s & Speedup & \#MAT & Tok./s & Speedup & \#MAT & Tok./s & Speedup & \\
\midrule
\multirow{6}{*}{\shortstack[l]{Qwen2.5\\7B-Instruct}} & Vanilla & 1.00 & 60.30 & 1.00$\times$ & 1.00 & 57.18 & 1.00$\times$ & 1.00 & 55.92 & 1.00$\times$ & 1.00$\times$ \\
 & PLD & 2.36 & 114.92 & 1.91$\times$ & 1.78 & 88.68 & 1.55$\times$ & 1.99 & 97.30 & 1.74$\times$ & 1.73$\times$ \\
 & TR & 3.04 & 131.58 & 2.18$\times$ & 2.75 & 114.56 & 2.00$\times$ & 2.65 & 112.34 & 2.01$\times$ & 2.06$\times$ \\
 & SAMD & 3.51 & 156.36 & 2.59$\times$ & 2.58 & 114.91 & 2.01$\times$ & 2.50 & 116.09 & 2.08$\times$ & 2.23$\times$ \\
 & ToolSpec & 4.82 & 184.48 & 3.06$\times$ & 4.23 & 158.00 & 2.76$\times$ & 3.93 & 148.73 & 2.66$\times$ & 2.83$\times$ \\
 & \textbf{OoO-Spec (Ours)} & \textbf{6.81} & \textbf{232.86} & \textbf{4.00$\times$} & \textbf{5.43} & \textbf{170.63} & \textbf{2.98$\times$} & \textbf{6.92} & \textbf{222.54} & \textbf{3.98$\times$} & \textbf{3.66$\times$} \\
\midrule
\multirow{6}{*}{\shortstack[l]{Qwen2.5\\14B-Instruct}} & Vanilla & 1.00 & 35.85 & 1.00$\times$ & 1.00 & 35.80 & 1.00$\times$ & 1.00 & 34.78 & 1.00$\times$ & 1.00$\times$ \\
 & PLD & 1.95 & 60.62 & 1.69$\times$ & 1.80 & 55.42 & 1.55$\times$ & 1.91 & 59.11 & 1.70$\times$ & 1.65$\times$ \\
 & TR & 2.87 & 75.40 & 2.10$\times$ & 2.67 & 69.23 & 1.93$\times$ & 2.42 & 61.54 & 1.77$\times$ & 1.94$\times$ \\
 & SAMD & 2.99 & 81.19 & 2.26$\times$ & 2.46 & 66.54 & 1.86$\times$ & 2.35 & 64.45 & 1.85$\times$ & 1.99$\times$ \\
 & ToolSpec & 4.36 & 103.71 & 2.89$\times$ & 3.96 & 92.21 & 2.58$\times$ & 3.54 & 85.15 & 2.45$\times$ & 2.64$\times$ \\
 & \textbf{OoO-Spec (Ours)} & \textbf{6.73} & \textbf{134.84} & \textbf{3.93$\times$} & \textbf{5.85} & \textbf{112.37} & \textbf{3.14$\times$} & \textbf{6.49} & \textbf{126.88} & \textbf{3.65$\times$} & \textbf{3.57$\times$} \\
\midrule
\multirow{6}{*}{\shortstack[l]{Llama-3.2\\3B-Instruct}} & Vanilla & 1.00 & 61.87 & 1.00$\times$ & 1.00 & 61.52 & 1.00$\times$ & 1.00 & 63.65 & 1.00$\times$ & 1.00$\times$ \\
 & PLD & 2.20 & 114.89 & 1.86$\times$ & 1.70 & 86.85 & 1.41$\times$ & 2.34 & 126.47 & 1.99$\times$ & 1.75$\times$ \\
 & TR & 3.10 & 136.78 & 2.21$\times$ & 2.88 & 121.34 & 1.97$\times$ & 2.92 & 131.38 & 2.06$\times$ & 2.08$\times$ \\
 & SAMD & 3.13 & 145.32 & 2.35$\times$ & 2.61 & 118.55 & 1.93$\times$ & 3.21 & 159.34 & 2.53$\times$ & 2.27$\times$ \\
 & ToolSpec & 5.24 & 201.83 & 3.26$\times$ & 3.48 & 136.94 & 2.23$\times$ & 3.27 & 140.42 & 2.21$\times$ & 2.57$\times$ \\
 & \textbf{OoO-Spec (Ours)} & \textbf{6.91} & \textbf{249.75} & \textbf{4.04$\times$} & \textbf{4.12} & \textbf{151.19} & \textbf{2.46$\times$} & \textbf{3.99} & \textbf{164.25} & \textbf{2.58$\times$} & \textbf{3.02$\times$} \\
\midrule
\multirow{9}{*}{\shortstack[l]{Llama-3.1\\8B-Instruct}} & Vanilla & 1.00 & 55.41 & 1.00$\times$ & 1.00 & 49.23 & 1.00$\times$ & 1.00 & 49.22 & 1.00$\times$ & 1.00$\times$ \\
 & PLD & 2.25 & 96.69 & 1.74$\times$ & 1.57 & 65.84 & 1.34$\times$ & 1.73 & 69.29 & 1.41$\times$ & 1.50$\times$ \\
 & TR & 3.37 & 125.74 & 2.27$\times$ & 3.03 & 103.09 & 2.09$\times$ & 2.85 & 99.20 & 2.02$\times$ & 2.13$\times$ \\
 & SAMD & 3.20 & 133.16 & 2.40$\times$ & 2.80 & 105.74 & 2.15$\times$ & 2.65 & 105.85 & 2.15$\times$ & 2.23$\times$ \\
 & EAGLE-3* & 3.07 & 79.26 & 1.43$\times$ & 3.13 & 67.99 & 1.38$\times$ & 3.09 & 74.85 & 1.52$\times$ & 1.44$\times$ \\
 & PARD-2* & 7.09 & 145.85 & 2.58$\times$ & \textbf{5.15} & 127.14 & 2.18$\times$ & \textbf{5.29} & 106.93 & 1.78$\times$ & 2.18$\times$ \\
 & DFlash* & 4.95 & 188.40 & 3.40$\times$ & 3.29 & 119.98 & 2.44$\times$ & 3.52 & 131.50 & 2.67$\times$ & 2.84$\times$ \\
 & ToolSpec & 5.50 & 186.87 & 3.37$\times$ & 2.78 & 88.41 & 1.80$\times$ & 2.36 & 79.98 & 1.62$\times$ & 2.26$\times$ \\
 & \textbf{OoO-Spec (Ours)} & \textbf{7.82} & \textbf{241.72} & \textbf{4.36$\times$} & 3.86 & \textbf{127.71} & \textbf{2.59$\times$} & 3.95 & \textbf{134.86} & \textbf{2.74$\times$} & \textbf{3.23$\times$} \\
\midrule
\multirow{7}{*}{\shortstack[l]{Qwen3\\4B-Instruct}} & Vanilla & 1.00 & 41.22 & 1.00$\times$ & 1.00 & 40.19 & 1.00$\times$ & 1.00 & 41.18 & 1.00$\times$ & 1.00$\times$ \\
 & PLD & 2.18 & 78.92 & 1.91$\times$ & 1.92 & 64.31 & 1.60$\times$ & 2.03 & 73.34 & 1.78$\times$ & 1.77$\times$ \\
 & TR & 3.45 & 110.70 & 2.69$\times$ & 3.36 & 104.29 & 2.59$\times$ & 3.01 & 98.37 & 2.39$\times$ & 2.56$\times$ \\
 & SAMD & 3.13 & 109.74 & 2.66$\times$ & 2.78 & 96.39 & 2.40$\times$ & 2.60 & 88.95 & 2.16$\times$ & 2.41$\times$ \\
 & ToolSpec & 5.84 & 170.40 & 4.13$\times$ & 4.65 & 131.72 & 3.28$\times$ & 4.12 & 128.12 & 3.11$\times$ & 3.51$\times$ \\
 & DFlash* & 6.23 & 180.34 & 4.38$\times$ & 5.31 & 151.61 & 3.77$\times$ & 6.35 & 175.51 & 4.26$\times$ & 4.14$\times$ \\
 & \textbf{OoO-Spec (Ours)} & \textbf{8.04} & \textbf{199.70} & \textbf{4.84$\times$} & \textbf{6.04} & \textbf{153.69} & \textbf{3.82$\times$} & \textbf{7.28} & \textbf{193.77} & \textbf{4.71$\times$} & \textbf{4.46$\times$} \\
\midrule
\multirow{8}{*}{\shortstack[l]{Qwen3\\8B-Instruct}} & Vanilla & 1.00 & 38.80 & 1.00$\times$ & 1.00 & 38.73 & 1.00$\times$ & 1.00 & 38.87 & 1.00$\times$ & 1.00$\times$ \\
 & PLD & 2.19 & 75.02 & 1.93$\times$ & 1.93 & 68.11 & 1.76$\times$ & 2.06 & 72.90 & 1.88$\times$ & 1.86$\times$ \\
 & TR & 3.58 & 112.72 & 2.91$\times$ & 3.53 & 110.15 & 2.84$\times$ & 3.19 & 96.00 & 2.47$\times$ & 2.74$\times$ \\
 & SAMD & 3.19 & 106.56 & 2.75$\times$ & 2.75 & 91.11 & 2.35$\times$ & 2.64 & 93.36 & 2.40$\times$ & 2.50$\times$ \\
 & ToolSpec & 5.87 & 163.97 & 4.23$\times$ & 4.85 & 128.08 & 3.31$\times$ & 4.15 & 128.04 & 3.29$\times$ & 3.61$\times$ \\
 & PARD-2* & 7.02 & 88.92 & 2.29$\times$ & 5.78 & 72.70 & 1.88$\times$ & 6.69 & 63.45 & 1.63$\times$ & 1.93$\times$ \\
 & DFlash* & 5.79 & 158.85 & 4.09$\times$ & 5.33 & 134.45 & 3.47$\times$ & 6.28 & 166.57 & 4.29$\times$ & 3.95$\times$ \\
 & \textbf{OoO-Spec (Ours)} & \textbf{8.35} & \textbf{207.09} & \textbf{5.34$\times$} & \textbf{6.85} & \textbf{153.68} & \textbf{3.97$\times$} & \textbf{7.42} & \textbf{189.95} & \textbf{4.89$\times$} & \textbf{4.73$\times$} \\
\midrule
\multirow{7}{*}{\shortstack[l]{Qwen3\\14B-Instruct}} & Vanilla & 1.00 & 34.97 & 1.00$\times$ & 1.00 & 33.14 & 1.00$\times$ & 1.00 & 34.65 & 1.00$\times$ & 1.00$\times$ \\
 & PLD & 2.18 & 65.42 & 1.87$\times$ & 1.87 & 56.75 & 1.71$\times$ & 1.88 & 58.58 & 1.69$\times$ & 1.76$\times$ \\
 & TR & 3.62 & 93.85 & 2.68$\times$ & 3.44 & 93.46 & 2.82$\times$ & 3.07 & 87.32 & 2.52$\times$ & 2.67$\times$ \\
 & SAMD & 3.13 & 95.21 & 2.72$\times$ & 2.94 & 87.44 & 2.64$\times$ & 2.77 & 83.49 & 2.41$\times$ & 2.59$\times$ \\
 & ToolSpec & 5.89 & 143.92 & 4.12$\times$ & 4.70 & 97.76 & 2.95$\times$ & 3.29 & 90.51 & 2.61$\times$ & 3.23$\times$ \\
 & PARD-2* & 6.75 & 120.96 & 3.46$\times$ & 5.27 & 109.42 & 3.30$\times$ & 6.46 & 56.24 & 1.62$\times$ & 2.79$\times$ \\
 & \textbf{OoO-Spec (Ours)} & \textbf{8.71} & \textbf{179.05} & \textbf{5.12$\times$} & \textbf{5.90} & \textbf{128.16} & \textbf{3.87$\times$} & \textbf{7.01} & \textbf{161.52} & \textbf{4.66$\times$} & \textbf{4.55$\times$} \\
\bottomrule
\end{tabular}%
}
\endgroup

\caption{Main comparison under greedy decoding; bold marks the best value per metric. \method{} always uses the same frozen split sidecar. $^{*}$Learned drafters use their faster measured official placement (EAGLE-3 split; PARD-2 and DFlash colocated); their rows appear only where an author-released checkpoint exists.}
\label{tab:primary}
\end{table*}

\subsection{Main Comparison}

\method{} is the fastest method in every cell it is measured in. Table~\ref{tab:primary} reports the complete matrix: across all seven targets and three benchmarks, \method{} leads in 21 of 21 cells, spans $2.46\times$--$5.34\times$ over live AR, and averages $3.89\times$ against $2.95\times$ for \toolspec{}. On the Qwen2.5 and Llama targets its overall speedups are $3.66\times$, $3.57\times$, $3.02\times$, and $3.23\times$, where the strongest prior methods reach $2.83\times$, $2.64\times$, $2.57\times$, and $2.84\times$; on Qwen3-4B, 8B, and 14B it reaches $4.46\times$, $4.73\times$, and $4.55\times$. Against each cell's strongest prior method, the per-target margin ranges from 7.7\% (Qwen3-4B) to 40.9\% (Qwen3-14B), averaging 23.4\%.

Accepted length alone does not decide these rankings. On Llama-3.1-8B, PARD-2 accepts more tokens per verification call than \method{} on ToolAlpaca (5.15 vs.\ 3.86) and BFCL (5.29 vs.\ 3.95), yet reaches only $2.18\times$ and $1.78\times$ where \method{} reaches $2.59\times$ and $2.74\times$. Drafting exposure, verifier cost, and runtime overhead all sit inside request latency; \method{} wins not by accepting the most tokens but by adding accepted length at almost no exposed cost.

\subsection{Transfer Without Target-Specific Training}
\label{sec:transfer}

Every \method{} number in this paper comes from one adapter, and the transfer happens along three axes at once. Across families and tokenizers: the sidecar is a Qwen3-0.6B trained on Qwen-generated semantic text, yet its largest single-benchmark speedup on a non-Qwen target is $4.36\times$ (Llama-3.1-8B, API-Bank), because the target re-tokenizes every hint and no Qwen token ID ever reaches a Llama model. Across target scales: the same 0.6B checkpoint serves 3B through 32B targets; it predicts schema slots, not target hidden states, so nothing about it refers to target width or depth. Across workloads: API-Bank, ToolAlpaca, and BFCL Java/JavaScript share the checkpoint, and BFCL was never represented in training at all.

The contrast with learned drafters is practical: every EAGLE-3, PARD-2, or DFlash result depends on a per-target training artifact, and a target without an author-released checkpoint cannot be served at all, whereas the sidecar asked nothing of any target but its tokenizer.

\subsection{Scaling to Larger Targets}

\begin{figure}[t]
\centering
\includegraphics[width=\columnwidth]{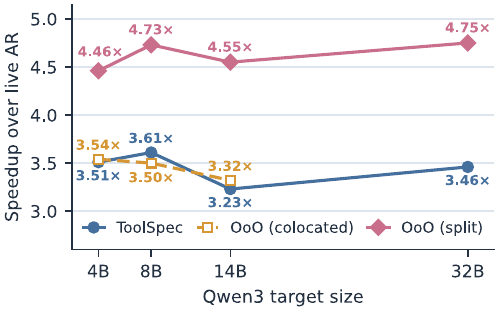}
\caption{Qwen3 target scaling with the frozen sidecar. Colocated \method{} stays close to \toolspec{}, while split placement sustains $4.46\times$--$4.75\times$; the colocated curve ends at 14B, where colocation no longer fits one GPU.}
\label{fig:qwen3-scale}
\end{figure}

Figure~\ref{fig:qwen3-scale} holds the sidecar fixed and grows the target from Qwen3-4B to 32B. \method{} averages $4.46\times$, $4.73\times$, $4.55\times$, and $4.75\times$ against \toolspec{}'s $3.51\times$, $3.61\times$, $3.23\times$, and $3.46\times$---relative improvements of 27.1\%, 31.0\%, 40.9\%, and 37.3\%, or 34.1\% on average. The direction is what the schedule predicts. The sidecar's work is a small structured object whose cost barely depends on the target; the target's per-step cost grows with its parameters; so every accepted semantic continuation displaces increasingly expensive verification steps. The same frozen sidecar also reaches an overall $4.58\times$ on Qwen2.5-32B.

\subsection{Placement Ablation}

Figure~\ref{fig:qwen3-scale} includes a placement ablation: the same frozen sidecar colocated on the target's GPU, measured in a separate run against its own AR reference. Colocated overall speedups are $3.54\times$, $3.50\times$, and $3.32\times$ on Qwen3-4B, 8B, and 14B---within $0.11\times$ of \toolspec{}---and Qwen3-32B cannot host both models on a single 80\,GB GPU. The ablation confirms the value of split placement: with the sidecar on its own device, the same frozen hints lift overall speedup from \toolspec{}-level to $4.46\times$--$4.75\times$, consistent with the critical-path analysis of Section~\ref{sec:system}.

\subsection{Sidecar Latency and Communication}
\label{sec:system}

Aggregated over 4{,}290 measured requests, the target path averages 309.5\,ms per request and the sidecar 85.0\,ms of wall time. The two run concurrently, so these numbers do not add: observed \method{} end-to-end latency is 311.9\,ms, only 2.4\,ms above the target path alone. Nearly all sidecar work hides behind target decoding.

The sidecar's 85.0\,ms is almost entirely generation (82.2\,ms). The remaining 2.79\,ms covers local transport, response processing, JSON decoding, and scheduling together, and thus bounds from above the full cost of delivering a hint. The semantic payload it delivers averages 85 bytes per request excluding protocol metadata. Communication in \method{} is therefore both infrequent (once per request) and small (tens of bytes), which is why device separation is cheap.

Timing also confirms that there is enough target work to overlap: 65.2\% of hints are ready before the first candidate construction, 98.6\% before the target finishes, and 94.6\% of hints are eventually used. The asynchronous design is not waiting on a rare fast path; on the large majority of requests the hint arrives early enough to matter and is actually consumed.

\section{Discussion}

\paragraph{Why split placement helps.}
\method{} benefits from a second GPU for two reasons. Semantic drafting is
asynchronous---the sidecar works from the request and schema without following
the target's current token position---and the separation needs little
communication: one compact semantic result per request rather than
target-conditioned states throughout decoding. Our system analysis
(Section~\ref{sec:system}) shows this payload and its response overhead are
small next to either model's execution, so most proposal work overlaps without
a high-bandwidth, step-synchronous device boundary.

\paragraph{Deployment implications.}
Because target and sidecar share no model dimensions, token IDs, or
device-local tensors, heterogeneous placement is possible in principle:
scarce high-memory accelerators reserved for the target, semantic drafting on
a lower-cost device, and independent scaling of both. One sidecar service
might also batch requests from multiple target replicas, amortizing its model
memory; evaluating heterogeneous hardware and multi-target serving is left to
future work.

\paragraph{On-device tool use.}
On-device agents are another natural setting: edge machines often pair an
integrated GPU with a discrete one, or two GPUs of very different capability,
behind interconnects far weaker than a server fabric. \method{} could keep the
target on the stronger device and draft semantics on the smaller one, since
the devices exchange a request-level semantic result rather than
step-synchronous hidden states; quantifying latency, energy, and concurrency
on edge hardware remains future work.

\section{Conclusion}

Tool calls are committed in textual order, but nothing forces their semantics to be computed that way. \method{} resolves a call's function and argument slots out of order on a small frozen sidecar, hands them to the target as re-tokenizable text, and lets \toolspec{}'s recurring candidate constructions absorb them whenever they are ready---while the target verifies and commits every token in order, exactly as before. One 0.6B sidecar, trained once, is the fastest method in all 21 ranked target--benchmark cells. Decoupling when a proposal is generated from when it is used gives a simple, transferable recipe for faster tool calling.

\bibliography{references}

\clearpage
\includepdf[pages=-]{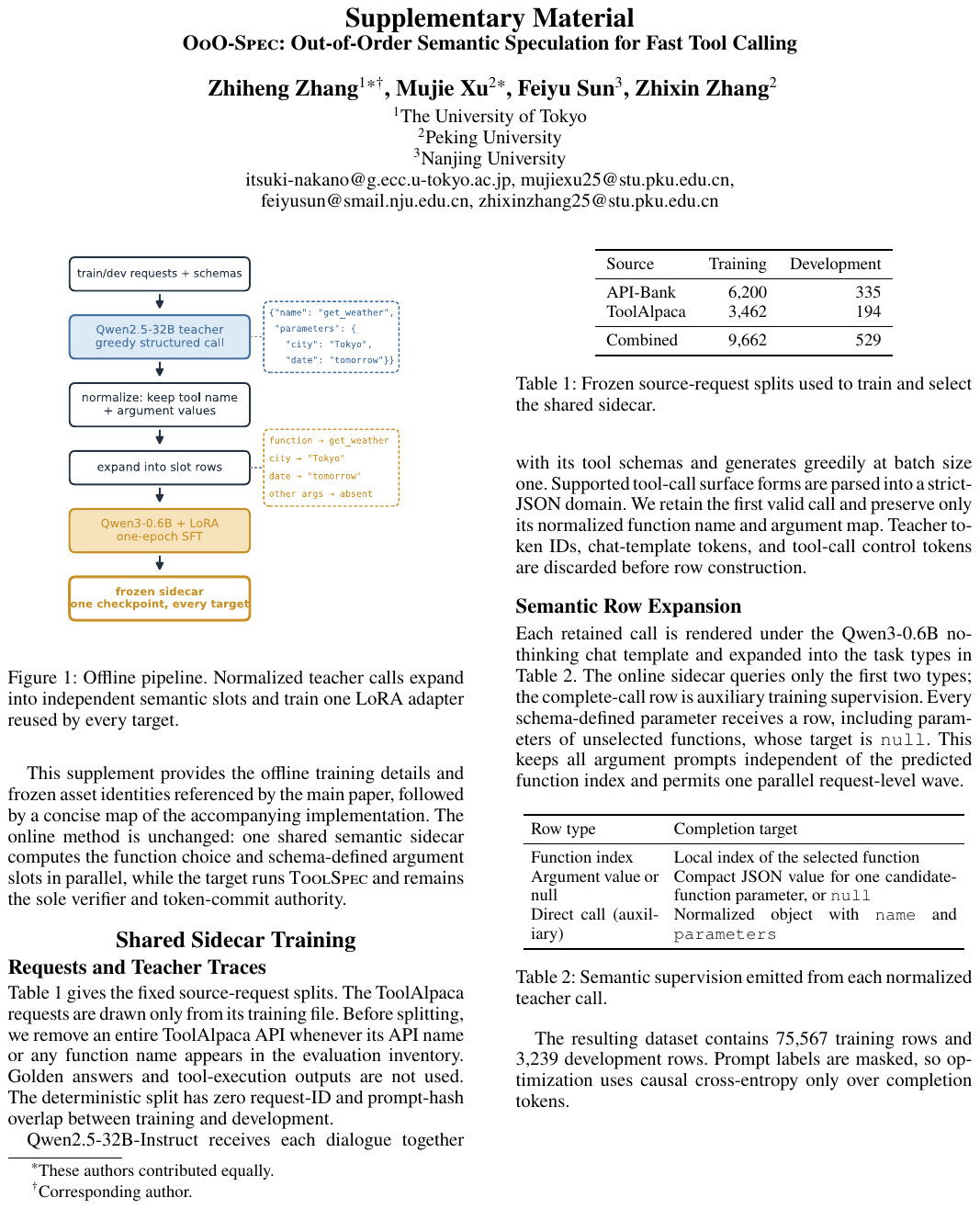}

\end{document}